\documentclass[runningheads]{llncs}

\usepackage[T1]{fontenc}
\usepackage{graphicx}
\usepackage{amsmath,amssymb}
\usepackage{booktabs}
\usepackage{multirow}
\usepackage{url}
\usepackage{hyperref}
\usepackage{float}
\hypersetup{hidelinks}
\begin{document}

\title{Group Resonance Network: Learnable Prototypes and Multi-Subject Resonance for EEG Emotion Recognition}
\titlerunning{Group Resonance Network for EEG Emotion Recognition}

\author{Renwei Meng\inst{1}\orcidID{0009-0001-6879-6629}}
\authorrunning{R. Meng}
\institute{Anhui University, Hefei, China\\
\email{R32314095@stu.ahu.edu.cn}}

\maketitle

\begin{abstract}
Electroencephalography (EEG)-based emotion recognition remains challenging in cross-subject settings due to severe inter-subject variability. Existing methods mainly learn subject-invariant features, but often under-exploit stimulus-locked group regularities shared across subjects. To address this issue, we propose the \emph{Group Resonance Network (GRN)}, which integrates individual EEG dynamics with offline group resonance modeling. GRN contains three components: an individual encoder for band-wise EEG features, a set of learnable group prototypes for prototype-induced resonance, and a multi-subject resonance branch that encodes PLV/coherence-based synchrony with a small reference set. A resonance-aware fusion module combines individual and group-level representations for final classification. Experiments on SEED and DEAP under both subject-dependent and leave-one-subject-out protocols show that GRN consistently outperforms competitive baselines, while additional analyses confirm the effects of prototype learning, PLV/coherence resonance, and leakage-safe reference construction.
\end{abstract}

\section{Introduction}
Electroencephalogram (EEG)-based emotion recognition is a key technique for affective brain-computer interfaces and human-machine interaction, and has been widely evaluated on public benchmarks such as DEAP, SEED, DREAMER, and AMIGOS \cite{koelstra2012deap,zheng2015seed,katsigiannis2018dreamer,mirandaCorrea2021amigos}. Although deep models, including CNNs, temporal-spatial networks, and graph-based methods, have achieved strong performance, they remain vulnerable to severe inter-subject variability in cross-subject settings \cite{lawhern2018eegnet,ding2023tsception,song2020dgcnn}.

Existing cross-subject methods mainly address this issue through transfer learning, domain adaptation, and subject-invariant representation learning \cite{pan2011tca,zheng2016personalizing,zheng2015transfercomponents,ganin2016dann,chai2016saae,bao2021tdann,li2021bidann,shen2023clisa}. However, most of them treat each subject as an isolated domain and under-exploit the \emph{shared, stimulus-locked} neural patterns that may emerge when different subjects experience the same affective stimulus. Neuroscience studies on inter-subject correlation and hyperscanning suggest that brain responses can exhibit measurable synchrony across individuals, often quantified by phase locking value (PLV) and coherence \cite{hasson2004isc,nastase2019isc,dmochowski2012correlated,nummenmaa2012emotions,montague2002hyperscanning,babiloni2014hyperscanning,hakim2023ibc,lachaux1999plv,nunez1997coherency,bastos2016connectivity}. These findings indicate that group-level reference structure may provide useful cues for robust affect decoding.

To this end, we propose the \emph{Group Resonance Network (GRN)} for EEG emotion recognition. GRN models \emph{offline} group resonance by computing PLV/coherence-based synchrony between a target subject and a small reference set, while simultaneously learning a set of \emph{learnable prototypes} that summarize group-consistent affective structure \cite{wang2021daspdnet,snell2017prototypical,caron2020swav}. Our contributions are three-fold: (i) we introduce stimulus-locked multi-subject resonance tensors for EEG emotion recognition; (ii) we combine learnable prototypes with resonance-aware fusion for group-level affect modeling; and (iii) we provide extensive analyses on component effects, reference alignment, statistical significance, and computational cost.

\section{Related Work}
\subsection{EEG Emotion Recognition and Cross-Subject Learning}
EEG emotion recognition has been widely studied on public benchmarks such as DEAP, SEED, DREAMER, and AMIGOS \cite{koelstra2012deap,zheng2015seed,katsigiannis2018dreamer,mirandaCorrea2021amigos}. Representative models include compact CNNs, temporal-spatial CNNs, graph neural networks, and recent Transformer-based variants \cite{lawhern2018eegnet,ding2023tsception,song2020dgcnn,song2023eegconformer,liu2024ertnet,si2024mactn}. To address cross-subject variability, prior work mainly adopts transfer learning, domain adaptation, adversarial alignment, or contrastive subject-invariant learning \cite{pan2011tca,zheng2015transfercomponents,zheng2016personalizing,chai2016saae,ganin2016dann,li2021bidann,bao2021tdann,shen2023clisa,he2020ea}. These methods reduce domain discrepancy, but rarely exploit group-level stimulus-locked structure explicitly.

\subsection{Inter-Subject Synchrony and Prototype Learning}
Common stimuli can induce synchronized neural responses across subjects, as reflected by inter-subject correlation and related shared-response analyses \cite{hasson2004isc,nastase2019isc,dmochowski2012correlated,dmochowski2014audience,nummenmaa2012emotions}. Hyperscanning research further measures inter-brain coupling using PLV and coherence \cite{montague2002hyperscanning,dumas2010interbrain,babiloni2014hyperscanning,hakim2023ibc}, grounded in electrophysiological connectivity analysis \cite{lachaux1999plv,nunez1997coherency,bastos2016connectivity}. Separately, prototype-based and metric learning methods represent latent structure with representative vectors \cite{snell2017prototypical,weinberger2009lmnn,caron2020swav}; in EEG, prototype learning has also been linked to domain adaptation on SPD representations \cite{wang2021daspdnet}. GRN connects these directions by integrating learnable prototypes with synchrony-derived offline resonance.

\section{Methodology}
\subsection{Overview and Definition of Resonance}
Given an EEG segment $X\in\mathbb{R}^{C\times T}$ and label $y$, GRN predicts the emotional state by combining an individual embedding, a prototype-induced group embedding, and a synchrony-based multi-subject embedding. In this paper, \emph{resonance} denotes stimulus-locked cross-subject alignment: two EEG responses are considered resonant if their phase or spectral coupling is high after being aligned by the onset and duration of the same affective stimulus. Thus, resonance is not a causal interaction as in online hyperscanning; it is an offline similarity/synchrony measure computed under shared stimulus timing.

\begin{figure}[t]
\centering
\includegraphics[width=\textwidth]{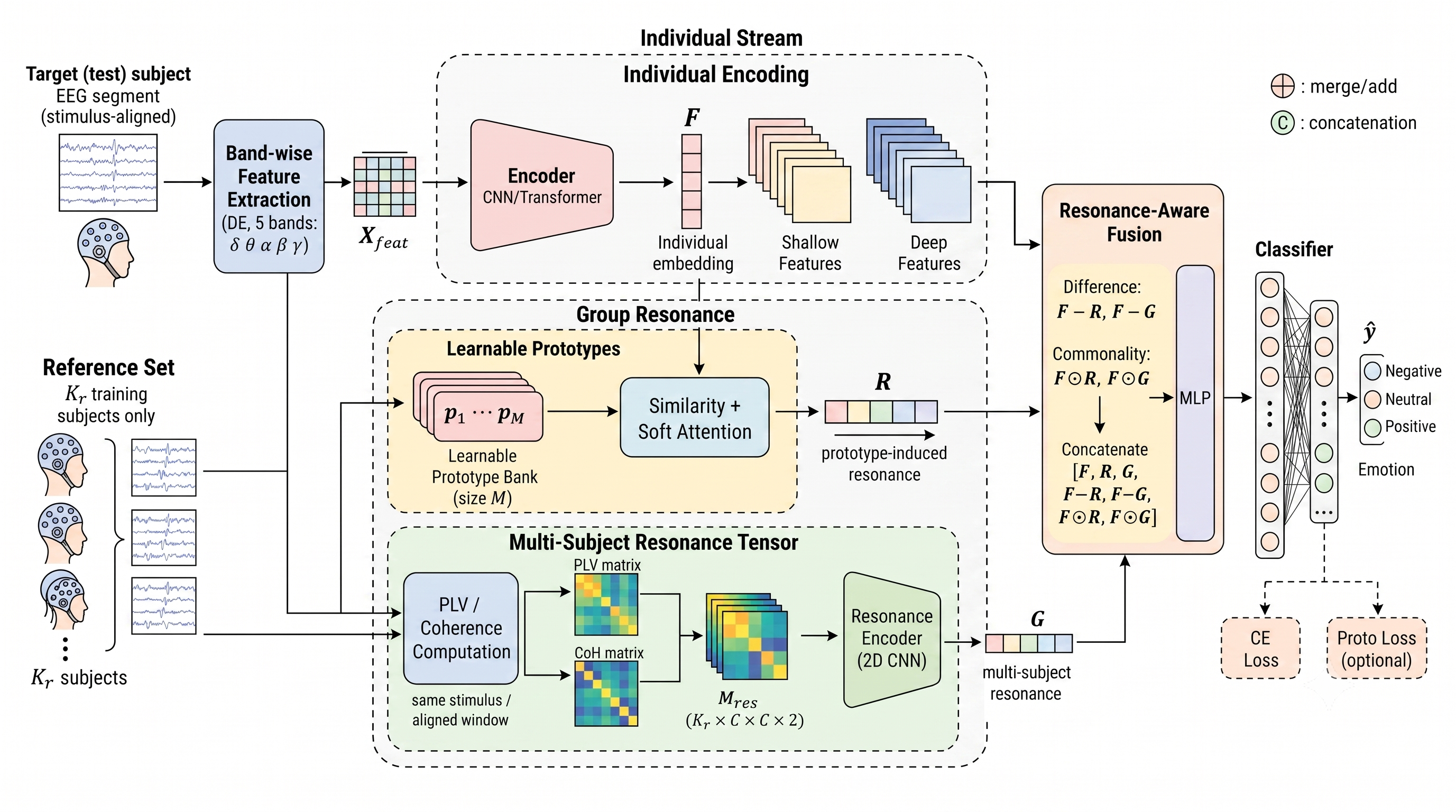}
\caption{Overall GRN pipeline. The model integrates an individual representation $F$, a prototype-induced group representation $R$, and a multi-subject resonance representation $G$ for emotion classification.}
\label{fig:overview}
\end{figure}

Figure~\ref{fig:overview} summarizes the architecture. The method contains four stages: individual encoding, learnable group prototype attention, multi-subject resonance tensor construction, and resonance-aware fusion.

\subsection{Individual Encoding}
We first extract band-wise features over five frequency bands ($\delta,\theta,\alpha,\beta,\gamma$), denoted as $X_{feat}$. An encoder $\mathrm{Enc}(\cdot)$ maps the feature tensor to a compact individual embedding:
\begin{equation}
F=\mathrm{Enc}(X_{feat})\in\mathbb{R}^{d}.
\end{equation}
The encoder is implemented as a lightweight convolutional-temporal backbone, but the proposed resonance modules are compatible with other EEG encoders such as graph networks or CNN-Transformer hybrids.

\subsection{Learnable Group Prototypes}
Instead of selecting a fixed prototype subject, GRN maintains $M$ trainable group prototypes:
\begin{equation}
\mathcal{P}=\{p_m\}_{m=1}^{M}, \quad p_m\in\mathbb{R}^{d}.
\end{equation}
For each sample, cosine similarity is computed between $F$ and each prototype:
\begin{equation}
s_m=\frac{F^{\top}p_m}{\|F\|_2\|p_m\|_2}.
\end{equation}
The attention weight over prototypes is obtained by temperature-scaled softmax:
\begin{equation}
\alpha_m=\frac{\exp(s_m/\tau)}{\sum_{j=1}^{M}\exp(s_j/\tau)}.
\end{equation}
The prototype-induced resonance embedding is then
\begin{equation}
R=\sum_{m=1}^{M}\alpha_m p_m.
\end{equation}
This design turns prototypes into trainable group anchors, allowing the model to learn latent affective patterns that are shared across subjects rather than depending on a single hand-selected reference.

\subsection{Multi-Subject Resonance Tensor}
For each training fold, a reference set $\mathcal{S}=\{X^{(k)}\}_{k=1}^{K_r}$ is sampled strictly from training subjects. In LOSO, the held-out test subject is never used as a reference. Moreover, resonance is computed only between samples aligned to the same stimulus timeline.

\begin{figure}[t]
\centering
\includegraphics[width=0.93\textwidth]{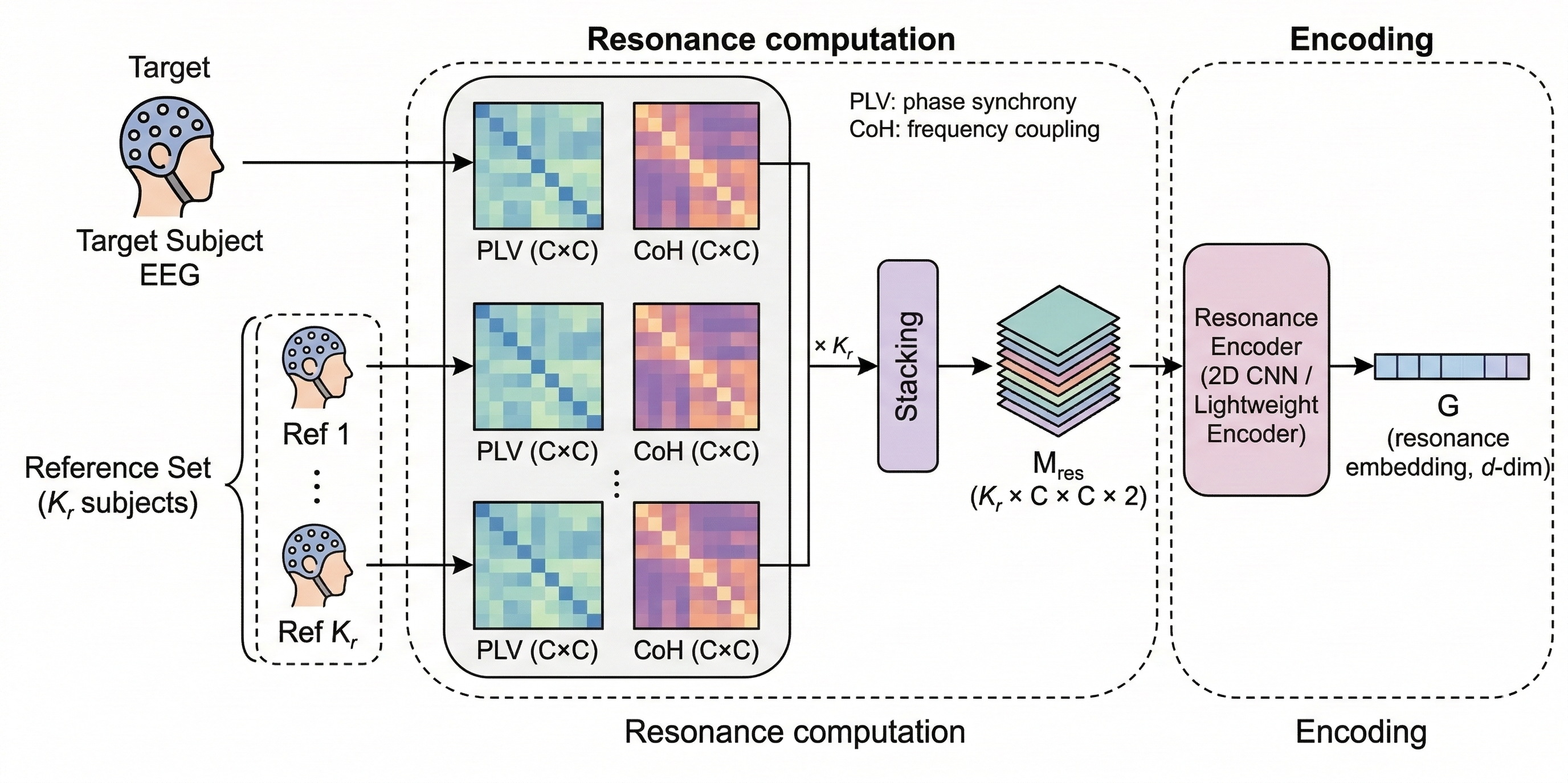}
\caption{Multi-subject resonance tensor construction. For each training-only reference subject, PLV and CoH matrices are computed under stimulus alignment and stacked into $\mathbf{M}$.}
\label{fig:res_tensor}
\end{figure}

For a target sample and reference $X^{(k)}$, PLV captures phase synchrony:
\begin{equation}
\mathrm{PLV}_{ij}^{(k)}=\left|\frac{1}{N}\sum_{n=1}^{N}e^{\mathrm{i}(\phi_i(n)-\phi_{j}^{(k)}(n))}\right|,
\end{equation}
where $\phi_i(n)$ and $\phi_j^{(k)}(n)$ are analytic phases estimated by the Hilbert transform. Coherence measures frequency-domain coupling:
\begin{equation}
\mathrm{CoH}_{ij}^{(k)}(f)=\frac{|P_{ij}^{(k)}(f)|^2}{P_{ii}(f)P_{jj}^{(k)}(f)},
\end{equation}
where $P_{ij}^{(k)}(f)$ is the cross-spectral density and $P_{ii}(f)$, $P_{jj}^{(k)}(f)$ are power spectral densities. PLV and band-averaged CoH are stacked into
\begin{equation}
\mathbf{M}\in\mathbb{R}^{K_r\times C\times C\times 2}.
\end{equation}
A shallow resonance encoder compresses $\mathbf{M}$ into
\begin{equation}
G=\mathrm{ResEnc}(\mathbf{M})\in\mathbb{R}^{d}.
\end{equation}
Figure~\ref{fig:res_tensor} illustrates the construction process.

\subsection{Resonance-Aware Fusion and Optimization}
GRN fuses $F$, $R$, and $G$ by explicitly modeling both difference and commonality:
\begin{equation}
D_R=F-R,\quad D_G=F-G,\quad C_R=F\odot R,\quad C_G=F\odot G.
\end{equation}
The final representation is
\begin{equation}
H=\mathrm{MLP}([F\|R\|G\|D_R\|D_G\|C_R\|C_G]),
\end{equation}
followed by a softmax classifier. The training loss is
\begin{equation}
\mathcal{L}=\mathcal{L}_{cls}+\lambda\mathcal{L}_{proto},\quad
\mathcal{L}_{proto}=\sum_{m=1}^{M}\alpha_m\|F-p_m\|_2^2.
\end{equation}
The prototype term encourages embeddings to remain close to the prototypes receiving high attention, stabilizing the learned group anchors. The additional resonance computation scales as $O(K_rC^2N)$ per sample before encoding; since $K_r$ is small and fixed, the overhead is bounded and can be precomputed for offline benchmarks.

\section{Experiments}
\subsection{Datasets, Protocols, and Implementation}
We evaluate GRN on SEED and DEAP. SEED is a three-class benchmark with Negative, Neutral, and Positive labels; DEAP is evaluated as binary Valence and Arousal classification by thresholding ratings at 5.0. We report subject-dependent (SD) results and subject-independent (SI) results using Leave-One-Subject-Out (LOSO). For every LOSO fold, references are sampled only from the training subjects. Default hyperparameters are $d=256$, $M=8$, $K_r=3$, batch size 64, Adam optimizer with learning rate $10^{-4}$ and weight decay $10^{-4}$, and early stopping with patience 10.

\subsection{Main Results}
Tables~\ref{tab:seed_main} and~\ref{tab:deap_main} report the main accuracy results. GRN achieves the best performance on both datasets without changing the standard evaluation protocols.

\begin{table}[H]
\caption{Accuracy (\%) on SEED under SD and SI protocols.}
\label{tab:seed_main}
\centering
\begin{tabular}{lcc}
\toprule
Method & SD & SI (LOSO) \\
\midrule
DGCNN & 90.51 $\pm$ 3.46 & 79.45 $\pm$ 10.11 \\
ST-DADGAT & 94.87 $\pm$ 2.51 & 83.59 $\pm$ 6.82 \\
FCAnet & 96.93 $\pm$ 2.28 & 84.16 $\pm$ 5.93 \\
LATN & -- & 86.15 $\pm$ 6.74 \\
DVIE-Net & 97.01 $\pm$ 1.93 & 86.34 $\pm$ 4.18 \\
\textbf{GRN} & \textbf{97.42 $\pm$ 1.61} & \textbf{87.90 $\pm$ 3.85} \\
\bottomrule
\end{tabular}
\end{table}

\begin{table}[H]
\caption{Accuracy (\%) on DEAP under SI protocol.}
\label{tab:deap_main}
\centering
\begin{tabular}{lcc}
\toprule
Method & Valence & Arousal \\
\midrule
DGCNN & 83.27 $\pm$ 4.30 & 82.36 $\pm$ 4.30 \\
ST-DADGAT & 86.61 $\pm$ 3.10 & 85.12 $\pm$ 5.80 \\
FCAnet & 88.92 $\pm$ 3.90 & 86.41 $\pm$ 3.40 \\
DVIE-Net & 89.27 $\pm$ 3.28 & 88.51 $\pm$ 3.76 \\
\textbf{GRN} & \textbf{90.35 $\pm$ 2.95} & \textbf{89.40 $\pm$ 3.20} \\
\bottomrule
\end{tabular}
\end{table}

Near-peer methods such as LATN and DVIE-Net also reduce inter-subject variability through transfer or prototype-style alignment, which explains their competitive performance. GRN differs by combining learned group anchors with explicit PLV/CoH synchrony, leading to consistent but moderate gains. A paired $t$-test across LOSO folds shows that GRN improves over DVIE-Net on SEED SI ($p=0.018$) and over the strongest DEAP baseline on Valence and Arousal ($p<0.05$). Table~\ref{tab:extra_metrics} further reports macro-F1 and AUC.

\begin{table}[H]
\caption{Additional SI metrics for GRN. AUC is macro one-vs-rest on SEED and binary on DEAP.}
\label{tab:extra_metrics}
\centering
\begin{tabular}{lccc}
\toprule
Task & Acc. (\%) & Macro-F1 (\%) & AUC \\
\midrule
SEED & 87.90 & 87.56 & 0.941 \\
DEAP-Valence & 90.35 & 90.12 & 0.922 \\
DEAP-Arousal & 89.40 & 89.08 & 0.911 \\
\bottomrule
\end{tabular}
\end{table}

\subsection{Ablation and Component Analysis}
Table~\ref{tab:ablation_seed} evaluates each component on SEED SI. Both group prototypes and multi-subject resonance improve the individual encoder, and removing the difference/commonality terms degrades performance. The PLV-only and CoH-only variants show that the two synchrony measures are complementary.

\begin{table}[H]
\caption{Ablation on SEED SI (LOSO).}
\label{tab:ablation_seed}
\centering
\begin{tabular}{lc}
\toprule
Variant & SI Acc. (\%) \\
\midrule
Individual only (remove $R,G$) & 84.60 \\
Capacity-matched individual encoder & 85.02 \\
+ Learnable prototypes only (add $R$, no $\mathbf{M}$) & 86.40 \\
+ Multi-subject resonance only (add $G$, no prototypes) & 86.85 \\
Full GRN w/o difference/commonality fusion & 86.92 \\
Full GRN w/o prototype regularizer & 87.35 \\
Full GRN with PLV only & 87.22 \\
Full GRN with CoH only & 86.98 \\
\textbf{Full GRN} & \textbf{87.90} \\
\bottomrule
\end{tabular}
\end{table}

Table~\ref{tab:ablation_deap} confirms that the component trend also holds on DEAP. Prototype attention analysis shows that learned prototypes form class-biased group anchors: the dominant prototype assignment purity reaches 0.81 on SEED, while the mean attention entropy decreases from 1.72 to 1.18 during training, indicating increasingly selective prototype usage.

\begin{table}[H]
\caption{Ablation on DEAP SI.}
\label{tab:ablation_deap}
\centering
\begin{tabular}{lcc}
\toprule
Variant & Valence & Arousal \\
\midrule
Individual only & 87.84 & 86.71 \\
+ Learnable prototypes only & 89.23 & 88.19 \\
+ Multi-subject resonance only & 89.41 & 88.55 \\
\textbf{Full GRN} & \textbf{90.35} & \textbf{89.40} \\
\bottomrule
\end{tabular}
\end{table}

\subsection{Stimulus Alignment, Sensitivity, and Cost}
Since SEED and DEAP assign labels through stimuli, the resonance branch must be interpreted as a stimulus-locked feature rather than a general-purpose hyperscanning signal. Table~\ref{tab:alignment} verifies this constraint: when references are stimulus-misaligned, the synchrony signal collapses toward the individual-only baseline. This supports that GRN exploits temporal synchrony rather than leakage from held-out subjects.

\begin{table}[H]
\caption{Reference alignment controls on SEED SI.}
\label{tab:alignment}
\centering
\begin{tabular}{lc}
\toprule
Reference condition & SI Acc. (\%) \\
\midrule
Aligned training references & \textbf{87.90} \\
Stimulus-misaligned training references & 84.83 \\
Random class-agnostic references & 84.65 \\
Label-shuffled reference pairing & 84.41 \\
Individual only & 84.60 \\
\bottomrule
\end{tabular}
\end{table}

Table~\ref{tab:sensitivity} shows that performance is stable across $K_r$ and $M$. The small variation suggests that GRN does not rely on a narrow hyperparameter choice; the best setting is used for all main experiments.

\begin{table}[H]
\caption{Sensitivity on SEED SI (LOSO).}
\label{tab:sensitivity}
\centering
\begin{tabular}{lccc}
\toprule
$K_r$ & 1 & 3 & 5 \\
\midrule
Acc. (\%) & 87.10 & \textbf{87.90} & 87.85 \\
\bottomrule
\end{tabular}
\quad
\begin{tabular}{lccc}
\toprule
$M$ & 4 & 8 & 12 \\
\midrule
Acc. (\%) & 87.45 & \textbf{87.90} & 87.88 \\
\bottomrule
\end{tabular}
\end{table}

\begin{table}[H]
\caption{Model size and runtime on SEED SI. Runtime is measured per batch.}
\label{tab:cost}
\centering
\begin{tabular}{lcc}
\toprule
Model & Params (M) & Runtime (ms) \\
\midrule
Individual encoder & 1.42 & 4.1 \\
Capacity-matched encoder & 1.86 & 4.7 \\
\textbf{GRN} & \textbf{1.89} & \textbf{5.6} \\
\bottomrule
\end{tabular}
\end{table}

\subsection{Visualization Analysis}
\begin{figure}[t]
\centering
\includegraphics[width=0.92\textwidth]{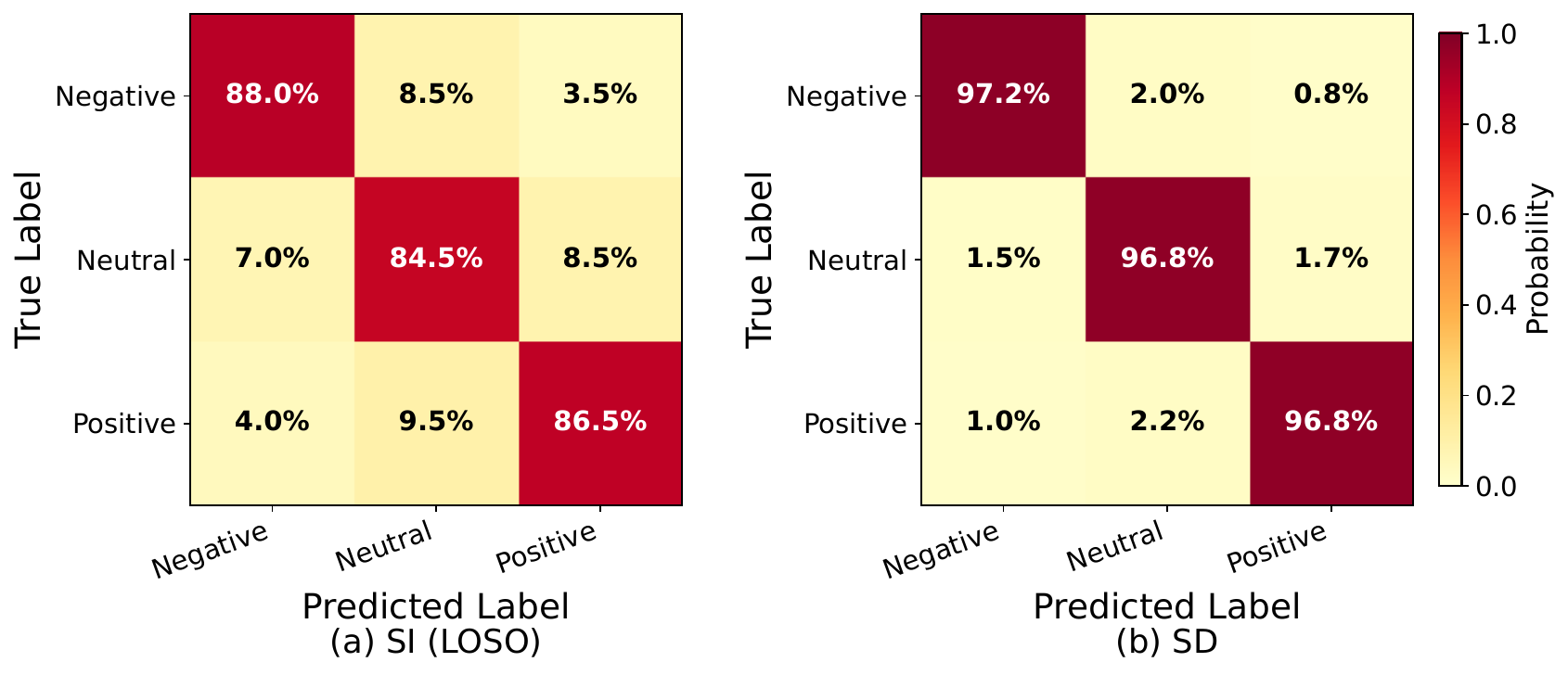}
\caption{Confusion matrices on SEED: (a) SI (LOSO) and (b) SD.}
\label{fig:confusion}
\end{figure}

\begin{figure}[t]
\centering
\includegraphics[width=0.92\textwidth]{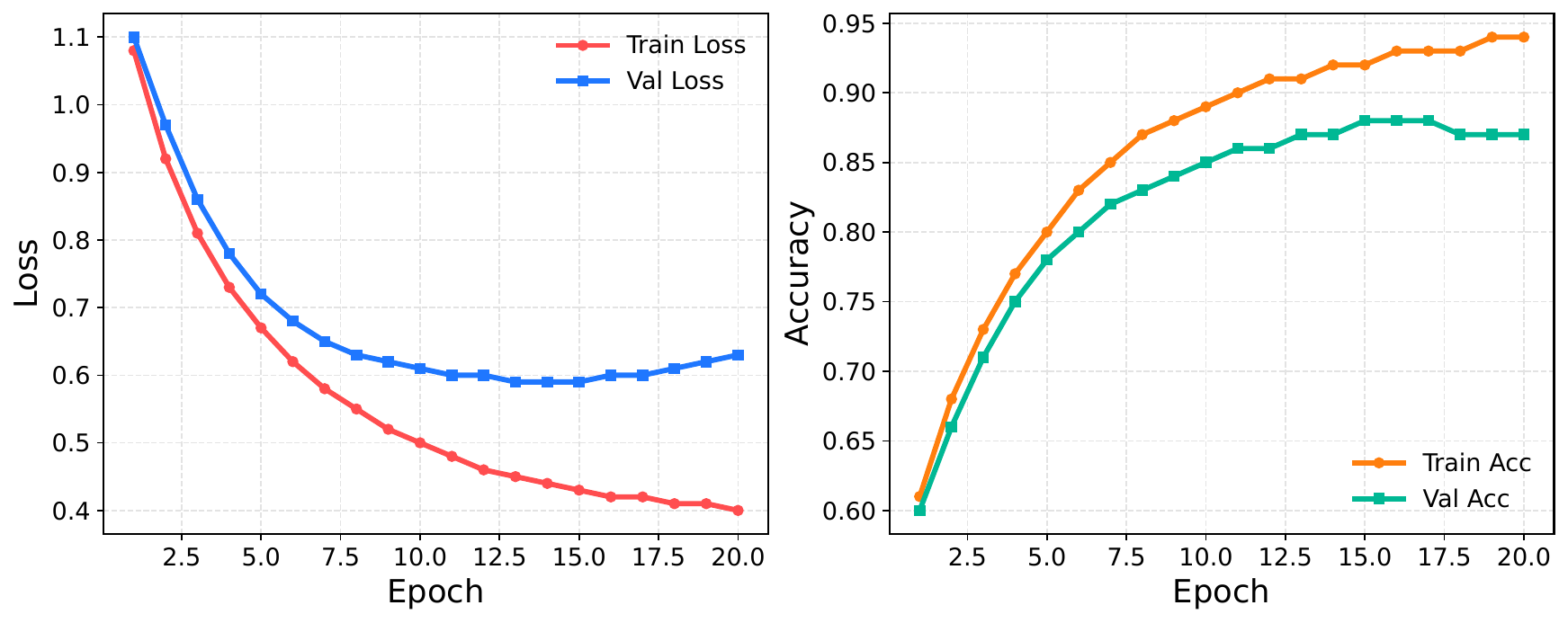}
\caption{Training/validation curves on SEED SI: loss and accuracy vs. epoch.}
\label{fig:curves}
\end{figure}

To further inspect class-wise behavior, Fig.~\ref{fig:confusion} reports the confusion matrices on SEED. GRN maintains balanced recognition under SI and becomes strongly diagonal under SD. The training curves in Fig.~\ref{fig:curves} show stable convergence without severe overfitting.

\section{Conclusion}
We presented GRN, a group-aware framework for EEG emotion recognition. GRN combines individual EEG encoding, learnable group prototypes, and PLV/CoH-based multi-subject resonance tensors under leakage-safe stimulus alignment. The resonance-aware fusion module jointly models subject-specific differences and group-level commonality. Experiments on SEED and DEAP show consistent improvements over competitive baselines, while additional analyses verify the contributions of prototype learning, PLV/CoH synchrony, fusion interactions, and reference alignment. The method is most suitable for offline affective benchmarks with shared stimulus timelines; extending it to finer-grained affect labels and weaker stimulus-label coupling is an important direction for future work.

\bibliographystyle{splncs04}
\bibliography{references}

\end{document}